\documentclass[10pt,twocolumn,letterpaper]{article}

\usepackage[margin=0.75in,columnsep=0.25in,top=0.85in,bottom=0.9in]{geometry}
\usepackage[T1]{fontenc}
\usepackage[utf8]{inputenc}
\usepackage{mathptmx}
\usepackage{courier}
\usepackage{microtype}
\usepackage{amsmath,amssymb}
\usepackage{graphicx}

\graphicspath{{figs/}}
\usepackage{booktabs,multirow,threeparttable}
\usepackage[font=small,labelfont=bf,skip=4pt]{caption}
\usepackage{xcolor}
\usepackage{enumitem}
\setlist{nosep,leftmargin=*}
\usepackage[backend=biber,style=ieee,sorting=nyt,maxcitenames=2,mincitenames=1,maxbibnames=99,giveninits=true,doi=false,isbn=false,url=false,eprint=false]{biblatex}
\usepackage{tikz}
\usetikzlibrary{positioning,arrows.meta,calc,fit,backgrounds}
\usepackage[colorlinks,linkcolor=blue!60!black,citecolor=blue!60!black,urlcolor=blue!60!black]{hyperref}
\usepackage{titlesec}
\titlespacing*{\section}{0pt}{9pt plus 2pt}{4pt}
\titlespacing*{\subsection}{0pt}{7pt plus 1pt}{3pt}
\titlespacing*{\paragraph}{0pt}{5pt}{6pt}
\titleformat*{\section}{\large\bfseries}
\titleformat*{\subsection}{\normalsize\bfseries}
\titleformat*{\paragraph}{\normalsize\bfseries}
\newcommand{\Rat}[1]{R@#1}
\newcommand{\best}[1]{\textbf{#1}}
\newcommand{\bestc}[1]{\underline{#1}}

\title{\vspace{-1.2em}%
\Large\bfseries An Invariant Tangent-Angle Descriptor and a Band U-Net for 2D Fragment Adjacency Prediction\vspace{-0.3em}}
\author{Guillaume Brouillette* \qquad
	Alain Goupil* \qquad Pierre-Olivier Parisé* \qquad Fadel Touré*\\[2pt]
	{\normalsize *Universit\'e du Qu\'ebec \`a Trois-Rivi\`eres, Trois-Rivi\`eres, Canada}\\
	{\small\texttt{\{guillaume.brouillette, alain.goupil, pierre-olivier.parise, fadel.toure\}@uqtr.ca}}\vspace{-0.6em}}
\date{}

\begin{document}
\maketitle

\begin{abstract}
Reassembling a broken artefact begins with deciding which fragments are adjacent. This paper addresses adjacency prediction for pairs of 2D fragments by analysing their contours. We propose a two-stage pipeline. In the first stage, each contour point is encoded using a tangent-angle profile, a descriptor that does not depend on rotation, translation or the choice of starting point. Pairs of profiles are then used to build an adjacency matrix, each entry computed either from a log-likelihood ratio or using a small 1D convolutional network. In the second stage, a U-Net classifies this matrix to predict whether the two fragments are adjacent. We compared our pipeline with two approaches from the literature using their own datasets. The results show that our approach outperforms them by reaching a Recall@10 of 0.823 versus 0.564 on the real dataset of PairingNet, and an accuracy of 0.979 versus 0.916 on Beaulac's dataset.
\end{abstract}

\section{Introduction}
\label{sec:intro}

The reassembly of broken flat objects, such as frescoes, ceramics or torn documents, is a recurrent problem in archaeology, cultural heritage preservation and forensic sciences \cite{Justino2006,Derech2021,Tsesmelis2024}. Most computational approaches to this problem start with a pairwise question: given two fragments, did they share a boundary? The answers for every pair of a collection form an adjacency graph, from which a global assembly can then be searched. The classical answer to the pairwise question is geometric. Since the two sides of a fracture have the same curve with opposite orientations, a signature of the contour based on curvature or on the turning angle should appear, reversed, on the contour of the adjacent fragment \cite{Wolfson1988,Arkin1991,Kong2001,Leitao2002,Zhu2006,Chen2008}. More recently, learning-based methods have been proposed, which rely on Siamese neural networks applied to fragment images \cite{Pirrone2019,Ostertag2020,Zhang2022}, on graph neural networks applied to contours and textures \cite{Zhou2024} or on Transformers applied to sequences of contour points \cite{Dongmo2025}. In these methods, the training labels are usually produced by a synthetic fragmentation generator.

\begin{figure*}[t]
	\centering
	\includegraphics[width=\linewidth]{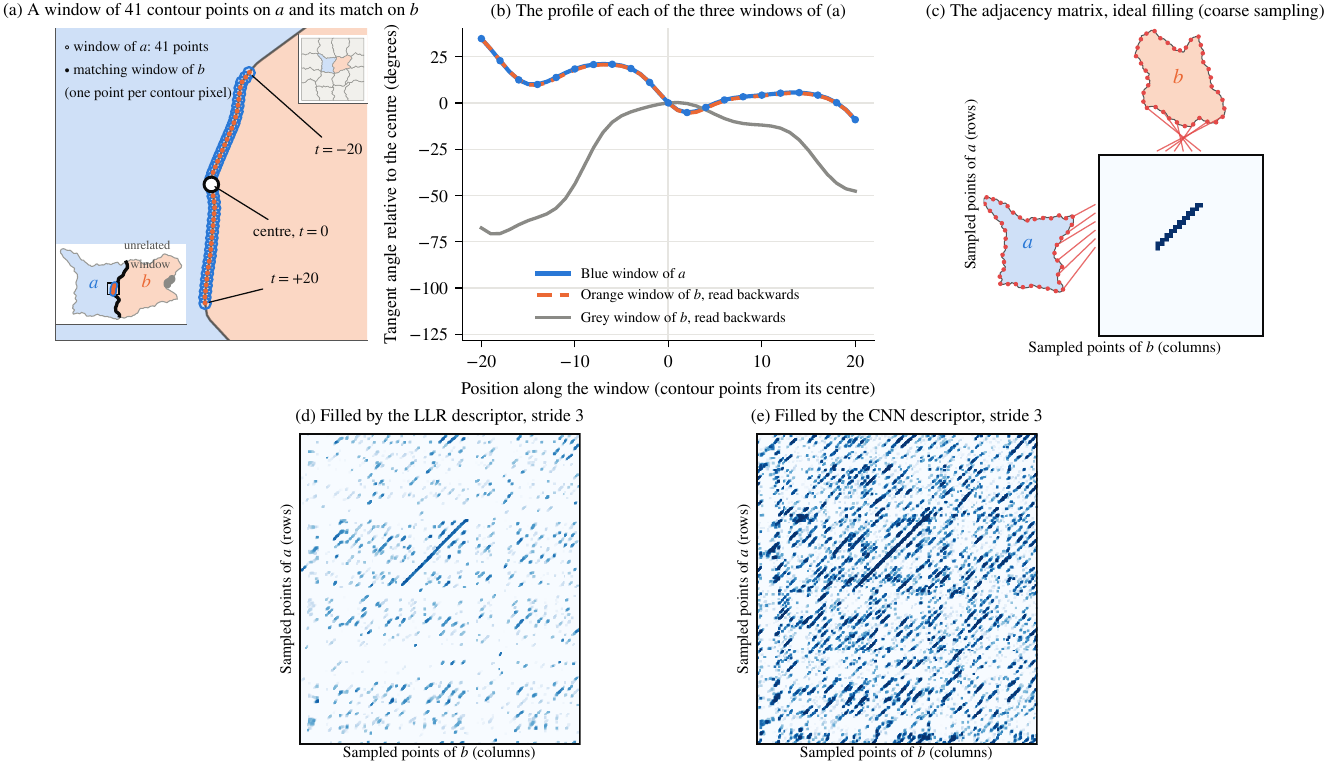}
	\caption{The pipeline on one synthetic pair (16-fragment image, validation split). (a) A window of $a$ centred on point $k$ (blue, the 41 contour points of the CNN descriptor) and the matching window of $b$ (orange), in a zoom of $52\times52$ pixels on the shared boundary. The insets show the whole pair, with the shared boundary in black and an unrelated window of $b$ in grey, and the image. (b) The profile of each window, namely the tangent angle relative to the angle at its centre, with the contours of $a$ and $b$ followed in opposite orientations. Dots mark the 21 samples seen by the CNN descriptor at one of its three scales. (c) The adjacency matrix: one row per sampled point of $a$ (red dots), one column per sampled point of $b$, so that a shared boundary is an anti-diagonal band. (d) The adjacency matrix filled by the LLR descriptor at stride 3. (e) The same adjacency matrix filled by the CNN descriptor.}
	\label{fig:overview}
\end{figure*}

The present work builds on the pipeline proposed by \textcite{Beaulac2023}. A trained network compares small neighbourhoods of the two contours and fills an adjacency matrix, in which a shared boundary appears as a band, and a second network classifies the pair of fragments from this matrix. We keep the adjacency matrix and modify what fills it and what reads it (see Figures \ref{fig:overview} and \ref{fig:pipeline}). Our contributions are the following.
\begin{itemize}
	\item We replace the network in \cite{Beaulac2023} applied to neighbourhoods, which we call the $C_8$ model, by a descriptor of the contour itself, based on the tangent angles along a window of contour points. This descriptor is invariant to rotation, translation and start point by construction. The comparison of two windows is either a log-likelihood ratio, the \emph{LLR descriptor}, or learned by a small one-dimensional convolutional network, the \emph{CNN descriptor}. Both descriptors are at least as accurate as the $C_8$ model for every classifier and every design of the synthetic benchmark (Table \ref{tab:synthetic}) and the LLR descriptor is more accurate on the dataset of \textcite{Beaulac2023} (Table \ref{tab:beaulac}). 
	\item We replace the ResNet18 that classifies the adjacency matrix by a U-Net trained to segment the band and to classify the pair of fragments at the same time. We also read the adjacency matrix with an \emph{alignment score}, which requires no training \cite{Smith1981,Chen2008}. The U-Net is the best classifier, or tied with the best, in 21 of the 24 cases of Table \ref{tab:synthetic}. Trained on synthetic data only, the pipeline transfers to the real dataset of PairingNet \cite{Zhou2024} with an AUC of 0.932 with the LLR descriptor and of 0.943 with the CNN descriptor (Table \ref{tab:transfer}).
	\item We evaluate these modifications on the synthetic benchmark of \cite{Beaulac2023}, regenerated with exact contours and point correspondences in four designs, on the published dataset of \cite{Beaulac2023}, whose original code, which is not public and was made available to us by \textcite{Dongmo2025}, we also re-run, and on the PairingNet dataset \cite{Zhou2024}. The accuracy of 72.45\% reported in \cite{Beaulac2023} is the consequence of a labelling error in the evaluation script, and the corrected pipeline of \cite{Beaulac2023} reaches 0.916 (0.939 with the roll augmentation of its code) against 0.979 for ours (Table \ref{tab:beaulac}). Under the pair-searching protocol of PairingNet, the CNN descriptor outperforms every method reported for the real dataset, including the published model, which also uses texture (Recall@10 of 0.823 against 0.564, Table \ref{tab:pairingnet}).
\end{itemize}

The remainder of the paper is structured as follows. In Section \ref{sec:related}, we review related work on curve matching and learned fragment matching. In Section \ref{sec:model}, we recall the pipeline of \textcite{Beaulac2023} and we describe the tangent descriptor, the U-Net and the alignment score. Section \ref{sec:results} presents the datasets, the experimental protocol and the results. Finally, Section \ref{sec:conclusion} concludes and discusses further work.

\section{Related work}
\label{sec:related}

\subsection{Curve matching approaches}
Matching fragments by means of an intrinsic signature of the contour curve is the classical approach to 2D reassembly. \textcite{Wolfson1988} match curvature strings of jigsaw pieces and \textcite{Arkin1991} introduce the distance between turning functions. \textcite{Kong2001}, \textcite{Leitao2002} and \textcite{McBride2003} match multiscale curvature signatures of sherds by dynamic programming on a matrix of pairwise costs. \textcite{Zhu2006} match segments of contours using the difference of their turning functions and \textcite{Ucoluk1999} use coordinate-independent signatures of crack curves which tolerate missing fragments. \textcite{Chen2008} apply the Smith--Waterman local alignment \cite{Smith1981} to sequences of turning angles for partial shape matching and read the high-scoring segments from the matrix of pairwise scores. Later systems combine such geometric matching with texture and with a global assembly step \cite{Zhang2014,Zhang2017,Derech2021,Cao2024,Shahar2025}. A survey of the field is given by \textcite{Lu2025survey}. These methods share a common pattern: a signature computed at each point of the contour, a matrix of pairwise scores between the points of the two contours and a reading of the diagonals of this matrix. The matrix is referred to as a cost matrix \cite{Kong2001,Leitao2002,McBride2003} or as an alignment matrix \cite{Chen2008}. The adjacency matrix of \textcite{Beaulac2023} has the same structure, with a trained neural network in place of the signature and a trained classifier in place of the dynamic programming. The LLR descriptor proposed in Section \ref{sec:descriptor} read with the alignment score of Section \ref{sec:classifier} belongs to this family. Compared with these methods, a likelihood ratio takes the place of the distance between the signatures of two contour points and a maximum-sum search along the anti-diagonals is a form of local alignment. The CNN descriptor follows the same idea, but learns the comparison from data instead of defining it explicitly.

\subsection{Learning-based approaches}
A first group of learning-based methods uses the image pixels to decide whether two fragments fit together. JigsawNet \cite{Le2019} applies a convolutional network to the image of a pair of fragments placed against each other. Siamese networks have been trained on images of papyrus \cite{Pirrone2019}, ostraca \cite{Ostertag2020,Ostertag2021} and oracle-bone fragments \cite{Zhang2022,He2024,ZhangZ2025}. Before deep learning, \textcite{Funkhouser2011} trained regression trees on hand-crafted properties of the contact region of fresco fragments and observed that these properties are the most predictive ones, which supports the idea of reading the band of the adjacency matrix.

A second group works directly from contours. \textcite{Beaulac2023} is our starting point and is described in Section \ref{sec:beaulac-model}. \textcite{Dongmo2025} replace the two stages of \textcite{Beaulac2023} by a Transformer encoder \cite{Vaswani2017} applied to patches of contour points with rotary positional encoding \cite{Su2024}, followed by a cross-attention module, a mean pooling and a pair classifier. They report an accuracy of 82\%. PairingNet \cite{Zhou2024} is the closest work to the learned part of our method. Its pair-matching module computes a matrix of pairwise scores between the points of two contours, referred to as a similarity matrix by the authors, from learned per-point features. These features combine a $7\times7$ binary patch of the contour image with texture features aggregated by a residual graph network over 8 neighbours on each side, and the matrix is supervised with the true point correspondences, as in our method. However, the per-point feature of PairingNet is an image patch, for which no invariance to rotation is claimed, and the matrix is only used to register a pair of fragments that has already been retrieved, by thresholding, morphological operations and RANSAC. The decision of whether two fragments are adjacent, called pair-searching by the authors, is taken from a separate global embedding trained with a contrastive loss, and the authors report that the contour-only variant of this embedding performs poorly (Recall@5 of 0.016). In our method, adjacency is read from the adjacency matrix itself. Other related works learn representations of curves for matching. WisePanda \cite{Zhu2026} learns an embedding of whole fracture curves of bamboo slips with a one-dimensional convolutional network and \textcite{Velich2023} learn invariant local signatures of planar curves by contrast for whole-shape matching. ReassembleNet \cite{Islam2025} uses the tangent angle and the curvature at learned keypoints as input features of a diffusion model for fresco reassembly. To our knowledge, no previous work learns a rotation-invariant tangent-angle descriptor of contour windows in order to build and read a complete contour-to-contour adjacency matrix.

\subsection{Datasets}
Synthetic fragmentation generators are used to produce training labels in large quantities. Examples are the recursive Perlin-noise shredder of \cite{Beaulac2023,Dongmo2025}, which we reuse, the Fourier-basis cuts of photographs used by PairingNet and the Voronoi partitions with erosion used in \cite{Shahar2025}. Real datasets with pairwise ground truth are less common. PairingNet \cite{Zhou2024} provides scanned fragments of hand-cut prints with annotated pairs and RePAIR \cite{Tsesmelis2024} provides eroded fresco fragments from Pompeii, on which assembly methods are compared by the precision of their neighbour predictions and by the error on the estimated position and orientation of each fragment
\cite{Islam2025,Scarpellini2024,Khoroshiltseva2024}. DAFNE \cite{Dondi2020} addresses a different task, namely the placement of fragments on a known fresco. Finally, solvers of square-grid puzzles, including those with eroded irregular cells \cite{Hosseini2023,Shahar2026}, do not consider the pairwise adjacency of free-form contours and are outside the scope of this paper.

\section{Methods}
\label{sec:model}

\subsection[The adjacency-matrix pipeline of Beaulac]{The adjacency-matrix pipeline of \citeauthor{Beaulac2023}}
\label{sec:beaulac-model}
In \cite{Beaulac2023}, a fragment is represented by its closed contour $c=(p_0,\dots,p_{L-1})$, with one point $p_t=(x_t,y_t), 0\leq t\leq L-1$ per pixel, oriented counter-clockwise. Since the contour is closed, the indices are taken modulo $L$. The pipeline of \cite{Beaulac2023} has two stages. The first stage, which we call the \emph{descriptor}, is a function $M(w_1,w_2)\in[0,1]$ that decides whether the \emph{neighbourhoods} $w_1$ and $w_2$ of two fragments are centred on the same point of a shared boundary. A neighbourhood, a term introduced in \textcite{Beaulac2023}, is a square crop centred on a contour point of the binary image of a fragment, in which the pixels have value 1 inside the fragment and 0 outside. In \cite{Beaulac2023}, this descriptor, which we call the \emph{$C_8$ model}, is a Siamese network. Each neighbourhood is embedded into $\mathbb{R}^{64}$ by a $C_8$-equivariant steerable convolutional network \cite{Weiler2019} made of six convolution layers with 24, 48, 48, 96, 96 and 64 regular fields and kernels of size 7, 5, 5, 5, 5 and 5, with antialiased pooling, a pooling over the group and a global average. The two embeddings are then concatenated and a two-layer head maps the result to a logit, that is, a real number whose sigmoid is $M (w_1, w_2)$.
The Siamese network is trained with the binary cross-entropy on pairs of neighbourhoods centred on common pixels of adjacent fragments, which are the positive examples, and on random points of two fragments, which are the negative examples. A random displacement of at most 3 pixels is applied to the centres, each neighbourhood is randomly rotated and the two neighbourhoods are randomly swapped. The size of the neighbourhoods retained by \textcite{Beaulac2023} is 29 pixels.

The \emph{adjacency matrix} $A$ of two fragments $a$ and $b$ is defined by $A_{kl}=M(w_a(ks),\,w_b(ls))$, where $w_a(t)$ is the neighbourhood centred on point $t$ of the contour of $a$ and the contour points are sampled with a stride $s=3$. Row $k$ of $A$ thus corresponds to the contour point of index $ks$ of $a$ and column $l$ corresponds to the contour point of index $ls$ of $b$. We denote by $n_a$ the number of rows and by $n_b$ the number of columns of $A$. If $L_a$ and $L_b$ are the numbers of points on the contours of $a$ and $b$, then $n_a = \lceil L_a/s \rceil$ and $n_b = \lceil L_b/s \rceil$, so that $A \in [0,1]^{n_a \times n_b}$. Since the encoder is a Siamese network, each neighbourhood is embedded only once and the head is applied to all $n_a n_b$ pairs of embeddings. When $a$ and $b$ share a boundary, their contours have opposite orientations along the boundary. Hence, the adjacency matrix contains what we call a \emph{band}, namely a sequence of consecutive high-valued entries along an anti-diagonal, that is, a line of entries $(k,l)$ with constant value $k+l$ (see Figure \ref{fig:overview}c). The entry $(0,0)$ is in the top-left corner of the adjacency matrix $A$. The second stage, which we call the \emph{classifier}, is a ResNet18 applied to the adjacency matrix after resizing it to a fixed square (224 pixels in \textcite{Beaulac2023}, 128 pixels in ours, with bilinear interpolation in both cases). It is trained on adjacency matrices of adjacent and non-adjacent pairs and outputs the adjacency logit. Note that the code of \textcite{Beaulac2023} applies a random cyclic roll to the rows and to the columns of the adjacency matrix during training. This \emph{roll augmentation} is appropriate, since the start point of each contour is arbitrary, but it is not described in \cite{Beaulac2023}. We therefore train the classifier both without and with the roll augmentation and report both results as ResNet18 and ResNet18 + roll, respectively.

Two fragments are labelled adjacent when the length of their shared boundary is at least $\beta=10\%$ of the perimeter of both fragments. This is the rule of the generator of the synthetic benchmark (Section \ref{sec:data}) as well as the rule used in \cite{Dongmo2025}. Pairs of fragments sharing a shorter boundary are labelled non-adjacent and we refer to them as \emph{near misses}.

\subsection{The tangent-angle descriptor}
\label{sec:descriptor}
\begin{figure*}[t]
\centering
\resizebox{0.96\linewidth}{!}{%
\begin{tikzpicture}[font=\small, x=1mm, y=1mm,
  box/.style={draw=black!60, rounded corners=2pt, align=center, inner sep=3pt, fill=white, text width=30mm, minimum height=12mm},
  ours/.style={box, draw=black!75, thick},
  ref/.style={box, draw=black!50, dashed, fill=black!3},
  stage/.style={draw=black!40, rounded corners=4pt, inner sep=5pt, fill=black!4},
  stagelabel/.style={font=\small\bfseries, inner sep=2pt},
  arr/.style={-{Latex[length=2mm]}, thick, draw=black!70}]
  \def\rowA{0}\def\rowB{-15}\def\rowC{-30}
  \node[ours] (llr) at (82,\rowA) {\textbf{LLR descriptor}\\Equation \eqref{eq:Allr}};
  \node[ours] (cnn) at (82,\rowB) {\textbf{CNN descriptor}\\Equation \eqref{eq:cell}};
  \node[ref]  (c8)  at (82,\rowC) {$C_8$ model\\{}\cite{Beaulac2023}, reference};
  \node[ours, text width=25mm, minimum height=27mm] (prof) at (44,{(\rowA+\rowB)/2}) {Windows and\\their profiles\\Equation \eqref{eq:profile}};
  \node[ref,  text width=25mm] (nbh) at (44,\rowC) {Neighbourhoods};
  \node[box, fill=blue!8,    text width=21mm] (ca) at (8,{\rowB+7.5}) {Contour of $a$};
  \node[box, fill=orange!10, text width=21mm] (cb) at (8,{\rowB-7.5}) {Contour of $b$,\\reversed};
  \node[box, text width=26mm, minimum height=27mm] (mat) at (120,\rowB) {\textbf{Adjacency}\\\textbf{matrix}\\$A\in[0,1]^{n_a\times n_b}$\\Figure \ref{fig:overview}c--e};
  \node[ours] (unet) at (158,\rowA) {\textbf{U-Net}\\ Adjacency logit};
  \node[ours] (kad)  at (158,\rowB) {\textbf{Alignment score}\\ No training};
  \node[ref]  (res)  at (158,\rowC) {ResNet18\\{}\cite{Beaulac2023}, reference};
  \node[ours, text width=20mm] (out) at (192,\rowB) {Adjacency decision};
  \begin{scope}[on background layer]
    \node[stage, fit=(prof)(nbh)(llr)(cnn)(c8)] (s1) {};
    \node[stage, fit=(unet)(kad)(res)] (s2) {};
  \end{scope}
  \node[stagelabel, anchor=south] at (s1.north) {Descriptor (Section \ref{sec:descriptor})};
  \node[stagelabel, anchor=south] at (s2.north) {Classifier (Section \ref{sec:classifier})};
    \coordinate (j1) at ($(ca.east) + (3,0)$);%
  \coordinate (j2) at ($(ca.east) + (7,0)$);%
  \draw[arr, -] (ca.east) -- (j1 |- ca.east) -- (j1 |- cb.east) -- (cb.east);
  \draw[arr, -] (j1 |- kad.west) -- (j2 |- kad.west);
  \draw[arr] (j2 |- kad.west) -- (j2 |- prof.west) -- (prof.west);
  \draw[arr] (j2 |- kad.west) -- (j2 |- nbh.west) -- (nbh.west);
  \draw[arr] (prof.east) -- (prof.east |- llr.west) -- (llr.west);
  \draw[arr] (prof.east) -- (prof.east |- cnn.west) -- (cnn.west);
  \draw[arr] (nbh.east) -- (c8.west);
  \draw[arr] (llr.east) -- (mat.west);
  \draw[arr] (cnn.east) -- (mat.west);
  \draw[arr] (c8.east) -- (mat.west);
  \draw[arr] (mat.east) -- (unet.west);
  \draw[arr] (mat.east) -- (kad.west);
  \draw[arr] (mat.east) -- (res.west);
  \draw[arr] (unet.east) -- (out.west);
  \draw[arr] (kad.east) -- (out.west);
  \draw[arr] (res.east) -- (out.west);
\end{tikzpicture}
}
\caption{The two stages of the pipeline. The windows of the two contours are described by their profiles and compared by the LLR descriptor or by the CNN descriptor, which fill the adjacency matrix. The adjacency matrix is read by the U-Net or by the alignment score. Dashed: the $C_8$ model and the ResNet18 of \textcite{Beaulac2023}, which serve as references.}
\label{fig:pipeline}
\end{figure*}

If the contour of fragment $b$ is reversed, the tangent angles of the contours of fragments $a$ and $b$, along the shared boundary, differ by a constant which corresponds to their relative rotation (see Figure \ref{fig:overview}b). Hence, we describe a window of a contour by its profile, namely its tangent angles relative to the tangent angle at the centre point of the window.

\subsubsection{Tangent angle}
The points of a contour are first smoothed with a periodic Gaussian filter of standard deviation $\sigma$, \begin{equation}
	\tilde p_t=(\tilde x_t,\tilde y_t)=\sum_{j} g_\sigma(j)\,p_{t-j}
\end{equation}
with weights $g_\sigma(j)\propto e^{-j^2/(2\sigma^2)}$ summing to one and indices are taken modulo $L$, which reduces the staircase pattern of the pixel chain. The tangent angle at point $t$ is the direction of the central difference $\tilde{p}_{t+1}-\tilde{p}_{t-1}$ of the smoothed points,
\begin{equation}
\theta(t)=\operatorname{atan2}\big(\tilde y_{t+1}-\tilde y_{t-1},\;\tilde x_{t+1}-\tilde x_{t-1}\big).
\label{eq:tangent}
\end{equation}
A rotation of the fragment adds a constant to every angle $\theta(t)$, a change of the start point shifts the index $t$. Reversing the orientation of the contour turns every tangent by $\pi$. For a fragment $a$, we write $\theta_a(t)$ for the tangent angle at point $t$ of its contour and $z_a(t)=e^{i\theta_a(t)}$. The contour of $b$ with reversed orientation, denoted by $\bar b$, has the tangent angles $\theta_{\bar b}(t)=\theta_b(t)+\pi$ and is read with decreasing indices. A \emph{window} is a set of $2h+1$ sample contour points, spaced $\Delta$ points apart around a centre point and indexed by $t=-h,\dots,h$. The \emph{profile} of the window of fragment $a$ centred on point $k$ is the sequence
\begin{equation}
\varphi_a(k,t)=\theta_a(k+t\Delta)-\theta_a(k),\qquad t=-h,\dots,h,
\label{eq:profile}
\end{equation}
and the profile of the window of $\bar b$ centred on point $l$ is $\varphi_{\bar b}(l,t)=\theta_{\bar b}(l-t\Delta)-\theta_{\bar b}(l)$, which walks backwards along $b$. A profile is invariant to rotation, translation and start point and, along a shared boundary, the profiles of two fragments coincide.

\subsubsection{LLR descriptor}
The mean resultant length of the profile of a window,
\begin{equation}
R_a(k)=\Big|\tfrac{1}{2h+1}\textstyle\sum_{t=-h}^{h}z_a(k{+}t\Delta)\Big|,
\label{eq:R}
\end{equation}
and likewise $R_{\bar b}(l)$ with $z_{\bar b}(l-t\Delta)$, is close to 1 when the samples of the window lie on a straight line and decreases as the window bends. For the window of fragment $a$ centred on $k$ and the window of $\bar b$ centred on $l$, we define
\begin{align}
I(k,l)&=\big(1-R_a(k)\big)+\big(1-R_{\bar b}(l)\big),\notag\\
D(k,l)&=1-\Big|\tfrac{1}{2h+1}\textstyle\sum_{t=-h}^{h}z_a(k{+}t\Delta)\,z_{\bar b}(l{-}t\Delta)^{*}\Big|,
\label{eq:DI}
\end{align}
where $^{*}$ denotes the complex conjugate. The quantity $I$ measures how much the two windows bend, that is, how informative they are, and $D$ is one minus the mean resultant length of the differences of angles $\theta_a(k+t\Delta)-\theta_{\bar b}(l-t\Delta)$, so that $D\approx\tfrac12\operatorname{var}(\theta_a-\theta_{\bar b})$ when these differences are concentrated around one value. Both quantities depend only on the two profiles: subtracting a constant from the angles of a window leaves $R$ unchanged, and subtracting $\theta_a(k)-\theta_{\bar b}(l)$ from every difference $\theta_a(k+t\Delta)-\theta_{\bar b}(l-t\Delta)$ leaves their mean resultant length, and hence $D$, unchanged. Two unrelated windows give $D\approx I$, two windows of the same curve give $D\approx\varepsilon$, where $\varepsilon$ is the angular noise floor, whatever their shape, and two windows on straight lines give $D\approx I\approx0$, which is no evidence either way. Under the hypothesis that the two windows come from the same curve, $D$ is modelled as the square of a zero-mean Gaussian residual with variance $\varepsilon$. Under the hypothesis that they are unrelated, the variance of this residual is $I+\varepsilon$. The log-likelihood ratio of the two hypotheses is
\begin{equation}
W(k,l)=\tfrac12\Big[\log\tfrac{I+\varepsilon}{\varepsilon}-\tfrac{D}{\varepsilon}+\tfrac{D}{I+\varepsilon}\Big].
\label{eq:W}
\end{equation} The value of this ratio is large for adjacent windows that are informative and close to zero for windows of straight lines. The \emph{LLR descriptor} fills the adjacency matrix at the same stride $s=3$ with
\begin{equation}
A_{kl}=\operatorname{sigmoid}\big(\max(W(ks,ls),-2)\big),
\label{eq:Allr}
\end{equation}
with $h=4$, i.e.\ 9 samples, $\Delta=4$, $\sigma=2$, $\varepsilon=0.003$ and a clip value of $-2$ on the ratio. These five parameters were chosen on the datasets of PairingNet \cite{Zhou2024}, as detailed in Appendix \ref{sec:provenance}. The classifier is used without modification.

\subsubsection{CNN descriptor}
The LLR descriptor compares the two profiles sample by sample, with a fixed noise floor $\varepsilon$, and we show in Section \ref{sec:pairingnet} that it fails under a jitter of 1 pixel of the contour points. The \emph{CNN descriptor} keeps the same profile but learns the comparison from examples with a one-dimensional convolutional network (CNN), so that the comparison can tolerate the noise seen in training. The profile of a window of fragment $a$ centred on point $k$, with $h=10$ and $\Delta=2$, is encoded as the array $u_a(k)\in\mathbb{R}^{6\times 21}$ of the pairs $(\cos\varphi,\;\sin\varphi)$, with $\varphi=\varphi_a(k,t)$ from Equation \eqref{eq:profile}, for $t=-h,\dots,h$ and for the three smoothing scales $\sigma\in\{1,2,4\}$ of the tangent angle. The array $u_{\bar b}(l)$ of the window of $\bar b$ centred on $l$ is defined from $\varphi_{\bar{b}}(l,t)$ in the same way. The arrays inherit the invariance of the profiles and the three scales describe the curve at three resolutions. Each array is embedded by the CNN $f\colon\mathbb{R}^{6\times 21}\to\mathbb{R}^{32}$ made of three convolutions of width 64 with kernels of size 5, 5 and 3, the second one with a stride of 2, GELU activations and a two-layer head, for a total of 129\,249 parameters. The entry of the adjacency matrix at stride $s$ is the log-odds
\begin{equation}
\operatorname{logit}A_{kl}=\tfrac{2}{\sqrt{32}}\,\langle f(u_a(ks)),\,f(u_{\bar b}(ls))\rangle+\eta,
\label{eq:cell}
\end{equation}
where $\eta$ is a learned bias. The adjacency matrix of a pair of fragments is thus a single product of the two arrays of embeddings, which are computed once per fragment.

The CNN descriptor is trained on the point correspondences of adjacent pairs. A positive example is the pair of windows centred on two matched points, a hard negative example replaces the window of the true partner by a window shifted by at least 24 points along $b$, and the other windows of the batch serve as easy negative examples. For a batch of 256 positive examples, the loss is the mean binary cross-entropy of the diagonal of the matrix of logits of all pairs of windows of the batch, which contains the positive examples, plus half the mean binary cross-entropy of its off-diagonal entries and half that of the hard negatives. The training contours are kept in four copies, one of which is clean while the other three are perturbed by an independent Gaussian jitter of the points whose standard deviation is drawn uniformly between 0.16 and 0.8 pixel per copy. Each window is drawn from a random copy. The optimiser settings are given in Appendix \ref{sec:hyper}.

\subsection{Classification of adjacency matrices}
\label{sec:classifier}
\subsubsection{Band U-Net}\label{sec:bandUNet}
The reference classifier is the ResNet18 of \cite{Beaulac2023}, with one input channel and either a random or an ImageNet initialisation. The classifier that we propose, which we call the \emph{band U-Net}, is a small U-Net with three levels of 16, 32 and 64 channels and two $3\times3$ convolutions with batch normalisation per level, applied to the $128\times128$ adjacency matrix. It has two outputs: a \emph{band map}, which gives, for every entry of the adjacency matrix, the probability that it belongs to the band, and an adjacency logit computed by a linear layer from the channel-wise average of the deepest feature map, of size $32\times32\times64$. Its loss is the binary cross-entropy on the logit for every adjacency matrix, plus the binary cross-entropy with a positive weight of 20 and the Dice loss on the band map for the adjacency matrices of adjacent pairs, whose ground-truth band is built from the correspondences given by the generator. The roll augmentation is applied to the adjacency matrix and to the band together. The adjacency decision is taken from the logit, except for the transfer to the real dataset (Section \ref{sec:transfer}), where the maximum of the band map is used as the score, since it transfers better. 

\subsubsection{Alignment score}\label{sec:alignmentscore}
The \emph{alignment score} reads the band without any training. It is the largest sum of $A-\tau$, with $\tau=0.5$, over the sequences of consecutive entries along the anti-diagonals of the adjacency matrix tiled $2\times2$. The tiling allows a band to wrap around the arbitrary start point of either contour. A dilation of one entry along the rows gives some tolerance to the slope of the band and a smoothing over three entries is applied along each anti-diagonal. The computed sum is the adjacency score. This is Kadane's algorithm applied to each anti-diagonal and corresponds to the special case of a Smith--Waterman alignment without gaps \cite{Smith1981,Chen2008}. The CNN descriptor uses this score on entries spaced 4 points apart, with the log-odds of each entry clipped below at $-2$, implemented as a matrix product followed by a cumulative-sum scan over all the fragments of a split.

\subsubsection{Invariances and cost}
The band U-Net and the alignment score are invariant to the start point thanks to the roll augmentation and to the tiling, respectively. In comparison, the $C_8$ model is equivariant to the eight rotations of $C_8$ only and relies on augmentation for intermediate angles, and its adjacency matrices are invariant to the start point only through the roll augmentation of the classifier. In terms of cost, the $C_8$ model embeds $L/s$ neighbourhoods per fragment with six equivariant convolutions, whereas the CNN descriptor embeds the windows centred on the same points with three one-dimensional convolutions on inputs of size $6\times21$ and the LLR descriptor needs no network at all.

\section{Experiments and results}
\label{sec:results}

\subsection{Datasets}
\label{sec:data}
\subsubsection{Synthetic benchmark}
We re-implemented the generator of \cite{Beaulac2023}, which cuts a square figure recursively, each cut being perturbed by one-dimensional Perlin noise \cite{Perlin1985} with fixed parameters (Appendix \ref{sec:hyper}). We call \emph{image} the square figure that is cut, together with its fragments. Whereas the generator of \cite{Beaulac2023} works on binary images and traces the contours as chains of pixels, our generator works on polygons with real-valued coordinates. Every fragment is a polygon and each cut is an exact polygon overlay. The contour of a fragment is the boundary of its polygon, sampled counter-clockwise at points one pixel apart along this boundary, so that the contour points have real-valued coordinates rather than pixel positions. For every pair of fragments that share a boundary, the shared boundary is the exact intersection of the two polygon boundaries. Thus, the labels of the boundary points on each contour and the point-to-point correspondence follow. In its original mode, the generator reproduces the images stored by \textcite{Beaulac2023} exactly. The images have a size of $1000\times1000$ pixels and contain either 16 fragments, as in the work of \textcite{Beaulac2023}, or 26 fragments, as in \cite{Dongmo2025}. We refer to them as 16-fragment and 26-fragment images.

Four variants, which we call designs, are generated for 16-fragment and 26-fragment images. We call \emph{frame} the boundary of the square image, \emph{exterior} a fragment whose contour touches the frame and \emph{interior} a fragment whose contour does not. In the \emph{Original} design, the benchmark is generated as published, with the contour points dropped when they are lying on the frame. Hence, the exterior fragments have open contours. In the \emph{Frame-free} design, only the interior fragments are kept. To compensate, three times as many images are generated and, for the 16-fragment images, 24 fragments are cut instead of 16. In the \emph{Frame-adjacent} design, the frame points are kept and a straight frame which is part of two exterior fragments counts as a shared boundary since the two fragments can be placed against each other. In this setting, two exterior fragments can therefore be considered adjacent even if they have no shared boundary apart from their straight frame parts. In the \emph{Varied cuts} design, the images are frame-free and every cut draws its own noise statistics, namely its amplitude, its lacunarity, its number of octaves, its deviation from the chord and the positions of its end points (Appendix \ref{sec:hyper}), so that the boundaries of a given image do not all look alike. The training, validation and test images are generated from disjoint ranges of seeds, with 400, 100 and 100 images for the \emph{Original} and \emph{Frame-adjacent} designs and 1\,200, 300 and 300 images for the two frame-free designs.

\subsubsection[Dataset of Beaulac]{Dataset of \citeauthor{Beaulac2023}}
The images used by \textcite{Beaulac2023} are public, whereas his code is not. They have a size of $400\times400$ pixels, 16 fragments, a noise amplitude of 0.07, a lacunarity of 5 and a gain of 2.25, and they come with the labels of the author, namely 2\,488 adjacent pairs in the 100 training images and 373 in the 15 validation images. We run his code end to end on these images (see Section \ref{sec:beaulac}) and we also run our re-implementation on the same images with his negative examples, which are all taken from different images, and his matrix size of 224 pixels.

\subsubsection{PairingNet}
PairingNet \cite{Zhou2024} provides a \emph{real dataset} and a \emph{generated dataset}. The real dataset contains 320 scanned fragments of 34 hand-cut printed images and 202 annotated pairs, 199 of which can be scored, the three others having an inconsistent ground truth. The generated dataset is split into training, validation and test splits. Its test split contains 3\,279 fragments and 2\,370 pairs, that is, 5.4 million candidate pairs, and its training split (4\,098 fragments and 3\,654 adjacent pairs with point correspondences) and validation split (819 fragments, 137 pairs) are used only to train the CNN descriptor. The contours are rescaled by a factor of 0.4 when the descriptors are used alone and by a factor of 0.8 for the pipelines trained on 1000-pixel images, so that the number of points of a contour is comparable to that of the synthetic contours.

\subsection{Experimental protocol}
\label{sec:protocol}
\subsubsection{Pipelines}
The $C_8$ model is trained on 2\,400 pairs of neighbourhoods, namely 1\,200 positive pairs at true correspondences and 1\,200 negative pairs, half of which come from another image, and it is validated on 4\,000 pairs with independently rotated neighbourhoods (Appendix \ref{sec:hyper}). We call \emph{$C_8$ pipeline}, \emph{LLR pipeline} and \emph{CNN pipeline} the pipelines built on each descriptor, and \emph{tangent pipelines} the last two together. The classifier of a pipeline is the U-Net unless specified otherwise. The classifier is trained on 2\,400 adjacency matrices built from 300 images and validated on 800 adjacency matrices built from 100 images, with balanced labels and negative examples drawn from the same image, near misses included, at a stride of 3, each adjacency matrix being resized to $128\times128$ pixels. Each classifier is trained for 30 epochs and the epoch with the best validation accuracy is kept. The test numbers of Table \ref{tab:synthetic} are obtained by applying this checkpoint once to 800 adjacency matrices built from the 100 held-out test images. Every pipeline of Tables \ref{tab:synthetic} and \ref{tab:transfer} is trained with three seeds and we report the means. The seed changes the training pairs, the initialisation and the batches of the classifier and, for the $C_8$ pipelines, of the $C_8$ model. The LLR descriptor has no training and the CNN descriptor is trained on synthetic correspondences of the \emph{Varied cuts} design (Table \ref{tab:crosstrain}) and used with the contours without rescaling. It is the same for every seed and every design. The validation, test and transfer pairs are the same for every seed. For pair-searching (Section \ref{sec:pairingnet}), the CNN descriptor is trained on the training split of PairingNet and its checkpoint is selected on the validation split. The real dataset and the test split are scored once.

\subsubsection{Evaluation on PairingNet}
Two measurements are made on the datasets of PairingNet \cite{Zhou2024}. The first measurement is a zero-shot adjacency test, in which a pipeline trained on synthetic data scores the 199 annotated pairs of the real dataset against 199 random pairs of fragments cut from different source images. Two fragments of the same source image are never used as a negative example: the annotation of the real dataset is not stated to be exhaustive \cite{Zhou2024}, hence a pair of the same image that is not annotated could still be a pair of neighbours, whereas two fragments of different images cannot. We report the AUC in this case, since the threshold of a classifier does not transfer from one dataset to another. The second measurement is pair-searching, in which every fragment is scored against every other fragment and the pair-searching protocol of PairingNet \cite{Zhou2024} is applied. This protocol makes one query per annotated pair, from the source fragment to the target fragment, with the whole split as the pool of candidates, and counts a hit when the target is within the top $k$ candidates. We use the evaluation functions published by \textcite{Zhou2024}, which our own implementation matches exactly for the recall and to within 0.001 for the NDCG. The pairs that cannot be scored count as misses and ties are resolved against our method. The published numbers of PairingNet, of JigsawNet \cite{Le2019} and of the rule-based method of \cite{Zhang2014} are taken from Tables 2 to 4 of \cite{Zhou2024} and were not re-run.

\subsubsection{Metrics}
The accuracy at the threshold of 0.5 is our primary metric, as in \cite{Beaulac2023,Dongmo2025}. We also report the area under the ROC curve (AUC) and the AUC restricted to the interior pairs, in which both fragments are interior, so that the open contour of an exterior fragment cannot be used as a cue. Pair-searching is measured by the Recall@$k$ and the normalised discounted cumulative gain NDCG@$k$ of the protocol of PairingNet \cite{Zhou2024}.

\subsection[Results on the dataset of Beaulac]{Results on the dataset of \citeauthor{Beaulac2023}}
\label{sec:beaulac}
\begin{table}[t]
	\centering\small
	\caption{Accuracy and AUC on the dataset of \textcite{Beaulac2023}, validation split. Rows 1--3: the code of \textcite{Beaulac2023}, mean over three seeds (standard deviation at most 0.014). Rows 4--11: our re-implementation, mean over three seeds (standard deviation at most 0.009). + ImageNet: ImageNet initialisation. + roll: roll augmentation. The code of \textcite{Beaulac2023} reports the accuracy only. Bold: best of the column.}
	\label{tab:beaulac}
	\footnotesize\setlength{\tabcolsep}{2pt}
	\begin{tabular}{@{}llcc@{}}
	\toprule
	Descriptor & Classifier & Acc. & AUC \\
	\midrule
	Code of \textcite{Beaulac2023} & ResNet18, labels as written & 0.715 & n/a \\
	 & ResNet18, labels corrected & 0.916 & n/a \\
	 & ResNet18, labels corrected + roll & 0.939 & n/a \\
	\midrule
	$C_8$ model (ours) & ResNet18 & 0.944 & 0.980 \\
	 & ResNet18 + ImageNet & 0.954 & 0.982 \\
	 & ResNet18 + roll & 0.951 & 0.985 \\
	 & U-Net & 0.960 & 0.989 \\
	LLR descriptor & ResNet18 & 0.980 & 0.991 \\
	 & ResNet18 + ImageNet & \best{0.983} & \best{0.995} \\
	 & ResNet18 + roll & 0.979 & 0.993 \\
	 & U-Net & 0.979 & 0.992 \\
	\bottomrule
	\end{tabular}
\end{table}

\textcite{Beaulac2023} reports accuracies of 72.32\% and 72.45\% for the classifier, with a ResNet18 initialised at random and with ImageNet weights respectively. Since these values are far below those obtained with our re-implementation, we ran the code of \textcite{Beaulac2023} on the published dataset with three training seeds, over which the standard deviation of every accuracy of this paragraph is at most 0.014. The $C_8$ model reaches a validation accuracy of 0.872, which is close to the values of \textcite{Beaulac2023}, 0.875 to 0.885. For the classifier, we observed that the evaluation script of this code labels every row of the pair table of a split as adjacent. The training table contains only adjacent pairs, but the validation table of the published dataset contains the 373 adjacent pairs together with 373 fixed non-adjacent pairs. For each of these 746 rows, the script produces the adjacency matrix of the row, labelled adjacent, and the adjacency matrix of a random non-adjacent pair, labelled non-adjacent, hence 1\,492 adjacency matrices for the validation split. As a result, 373 of them carry a wrong label and no classifier can exceed an accuracy of $(373+746)/1\,492=75.0\%$. The training labels are not affected. With the labels as written, the classifier of the author gives 0.715. With the corrected labels, the same classifier gives 0.916 (0.906 to 0.932 over the three seeds) with the settings of \cite{Beaulac2023}, which do not mention any augmentation of the adjacency matrices, and 0.939 with the roll augmentation of the code. Larger neighbourhoods are reported in Appendix \ref{sec:beaulac63}. The value of 72\% is due to the evaluation script rather than to the method. In particular, the conclusion of \cite{Beaulac2023} that the classifier does not generalise should be reconsidered.

Table \ref{tab:beaulac} compares our re-implementation of the $C_8$ model and the LLR descriptor on the same images with the labels of \textcite{Beaulac2023}, on the 746 pairs of the validation table. Over three seeds, our re-implementation of the $C_8$ pipeline gives mean accuracies between 0.944 and 0.960, close to those of the corrected code of the author (0.906 to 0.950 over the three seeds, with and without the roll augmentation). With the LLR descriptor, every classifier reaches a mean accuracy between 0.979 and 0.983. The U-Net is the best classifier for the $C_8$ model and is within 0.004 of the best one for the LLR descriptor. The images of \textcite{Beaulac2023} have a size of $400\times400$ pixels, against $1000\times1000$ pixels for our benchmark, and their contours are pixel chains traced on the binary images of the fragments, with integer coordinates, rather than resampled polygons, which the smoothing of Equation \eqref{eq:tangent} must handle.

\subsection{Synthetic benchmark}
\label{sec:synthetic}
\begin{table*}[t]
	\centering\small
	\begin{threeparttable}
	\caption{Accuracy on the held-out test split of the synthetic benchmark, mean over three seeds. Standard deviation at most 0.025 for the accuracies of the tangent descriptors and 0.108 for those of the $C_8$ model, 0.004 and 0.091 for the interior-pair AUC. \emph{Int.}: interior-pair AUC of the U-Net. Bold: best of the design in each column.}
	\label{tab:synthetic}
	\footnotesize\setlength{\tabcolsep}{3.2pt}
	\begin{tabular}{@{}llccccc|cccc@{}}
	\toprule
	& & \multicolumn{5}{c|}{16-fragment images} & \multicolumn{4}{c}{26-fragment images} \\
	Design & Descriptor & ResNet18 & + ImageNet & + roll & U-Net & Int. & ResNet18 & + ImageNet & + roll & U-Net \\
	\midrule
	\multirow{3}{*}{\emph{Original}} & $C_8$ model & 0.916 & 0.970 & 0.980 & 0.996 & 1.000 & 0.800 & 0.914 & 0.938 & 0.975 \\
	  & LLR descriptor & \best{0.995} & \best{0.997} & 0.997 & \best{0.997} & 1.000 & \best{0.973} & \best{0.975} & \best{0.978} & 0.982 \\
	  & CNN descriptor & 0.991 & 0.993 & \best{0.998} & 0.996 & 1.000 & 0.969 & 0.974 & 0.975 & \best{0.983} \\
	\midrule
	\multirow{3}{*}{\emph{Frame-free}} & $C_8$ model & 0.592 & 0.864 & 0.905 & 0.962 & 0.993 & 0.599 & 0.871 & 0.925 & 0.958 \\
	  & LLR descriptor & \best{0.961} & \best{0.968} & \best{0.974} & \best{0.982} & 0.999 & \best{0.954} & \best{0.965} & \best{0.970} & \best{0.983} \\
	  & CNN descriptor & 0.945 & 0.948 & 0.962 & 0.974 & 0.994 & 0.947 & 0.951 & 0.959 & 0.976 \\
	\midrule
	\multirow{3}{*}{\emph{Frame-adjacent}} & $C_8$ model & 0.915 & 0.912 & 0.920 & 0.960 & 0.893 & 0.873 & 0.885 & 0.903 & 0.909 \\
	  & LLR descriptor & \best{0.992} & \best{0.994} & 0.993 & \best{0.995} & 1.000 & 0.928 & 0.948 & 0.945 & 0.951 \\
	  & CNN descriptor & \best{0.992} & 0.991 & \best{0.994} & 0.992 & 0.994 & \best{0.944} & \best{0.952} & \best{0.963} & \best{0.961} \\
	\midrule
	\multirow{3}{*}{\emph{Varied cuts}} & $C_8$ model & 0.590 & 0.712 & 0.818 & 0.873 & 0.937 & 0.570 & 0.661 & 0.810 & 0.869 \\
	  & LLR descriptor & 0.793 & 0.834 & 0.867 & 0.888 & 0.954 & 0.764 & 0.831 & 0.867 & 0.903 \\
	  & CNN descriptor & \best{0.887} & \best{0.913} & \best{0.928} & \best{0.938} & 0.987 & \best{0.892} & \best{0.911} & \best{0.945} & \best{0.958} \\
	\bottomrule
	\end{tabular}
	\end{threeparttable}
\end{table*}

\begin{figure*}[t]
	\centering
	\includegraphics[width=0.92\linewidth]{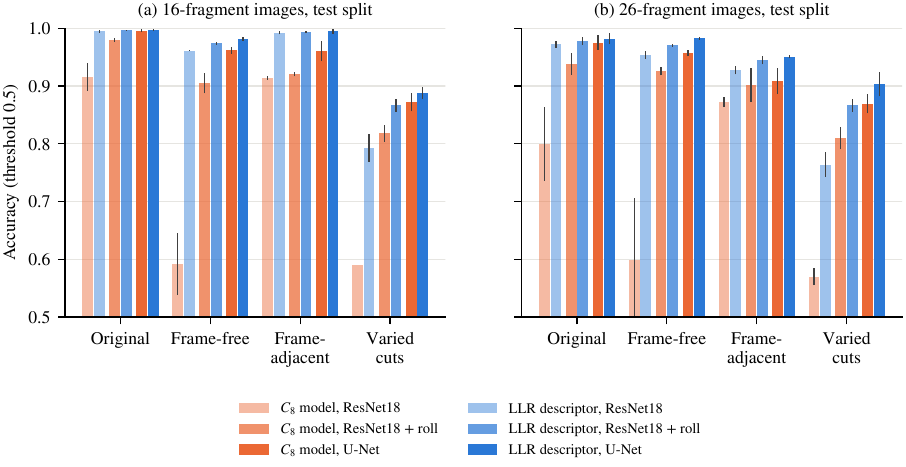}
	\caption{Accuracy on the synthetic benchmark (Table \ref{tab:synthetic}) by descriptor (orange: $C_8$ model, blue: LLR descriptor) and by classifier (light to dark: ResNet18, ResNet18 + roll, U-Net, the ImageNet variant being omitted), test split, mean over three seeds (error bars: one standard deviation). (a) 16-fragment images. (b) 26-fragment images.}
	\label{fig:synthetic}
\end{figure*}

Table \ref{tab:synthetic} and Figure \ref{fig:synthetic} give the complete comparison on the synthetic benchmark.

First, each of the two tangent descriptors performs as well as or better than the $C_8$ model in each of the 32 configurations, a \emph{configuration} being a choice of design, number of fragments per image and classifier. The difference is largest where the classifier is weakest. With the plain ResNet18 of \textcite{Beaulac2023}, the $C_8$ pipeline falls to 0.592 and 0.599 on the \emph{Frame-free} design and to 0.590 and 0.570 on the \emph{Varied cuts} design, whereas the LLR pipeline remains at 0.961 and 0.954 on the former and at 0.793 and 0.764 on the latter. The adjacency matrices produced by the $C_8$ model are noisier, since this model is correct on 82\% to 88\% of the rotated pairs of neighbourhoods (validation accuracy of the $C_8$ model, mean over three seeds, according to the design). A classifier trained from scratch on these adjacency matrices at a resolution of 128 pixels does not detect the band reliably. The adjacency matrices produced by the tangent descriptor are cleaner (see Figure \ref{fig:overview}d) and every classifier can read them. On the \emph{Original} design, part of the accuracy comes from the open contours of the exterior fragments, which exhibit a straight gap where the frame was. This is consistent with all three pipelines performing best on this design and with the plain ResNet18 of the $C_8$ pipeline losing 0.324 of accuracy on the 16-fragment images when the frame is removed, i.e.\ when changing from the \emph{Original} design to the \emph{Frame-free} design, while the LLR and CNN pipelines lose 0.034 and 0.046.

Second, for each descriptor, design and number of fragments, the U-Net is the best classifier, or tied with the best, in 21 of the 24 cases. The roll augmentation is at least as good as the ImageNet initialisation in 22 of them. The plain ResNet18 of \cite{Beaulac2023} is the worst in 22 of them, by a margin that depends on the descriptor (between 0.036 and 0.370 below the U-Net with the $C_8$ model, between 0.002 and 0.139 with the LLR descriptor and between 0.000 and 0.066 with the CNN descriptor). The interior-pair AUC of the U-Net lies between 0.937 and 1.000 for every design, except for the \emph{Frame-adjacent} design with the $C_8$ model (0.893, with seeds between 0.839 and 0.998). Thus, the pipelines do not rely on the frame to decide adjacency.

Third, the CNN descriptor trained on synthetic correspondences, used with the contours without rescaling, is close to the LLR descriptor on three designs and ahead of it on the fourth. On the \emph{Original}, \emph{Frame-adjacent} and \emph{Frame-free} designs, the two descriptors are within 0.020 of each other for every classifier and both numbers of fragments. On the \emph{Varied cuts} design, the CNN descriptor is ahead for every classifier, by 0.050 to 0.055 for the U-Net and by 0.094 to 0.128 for the plain ResNet18. Hence the learned comparison of Section \ref{sec:descriptor} seemingly brings its gain where the boundaries are the most varied, that is, where the fixed noise floor of the LLR descriptor is the least adequate. This design is also the one on which the CNN descriptor was trained, so part of this gain may come from the match between its training and test designs.

Fourth, the \emph{Varied cuts} design is the most difficult one. When every cut of an image has its own noise parameters, the accuracy decreases for every configuration (between 0.869 and 0.958 for the U-Net). Two likely causes are that smooth cuts of low amplitude give short informative boundaries and that the near misses of one image resemble the positive pairs of another. As shown in Section \ref{sec:transfer}, this is also the design that transfers best to the real dataset. Finally, the mean validation and test accuracies agree within 0.029 in every configuration, so the selection of the checkpoint on the validation split does not inflate the numbers of Table \ref{tab:synthetic}.

\subsection{Label rule and stride}
\begin{table}[t]
	\centering\small
	\caption{Validation accuracy of the four classifiers on the \emph{Original} design with the $C_8$ model, by label rule, input size and stride, mean over three seeds (standard deviation at most 0.024, except up to 0.074 for the plain ResNet18 with an input of 128 pixels). The 26-fragment images use the 10\% rule. + ImageNet: ImageNet initialisation. + roll: roll augmentation.}
	\label{tab:ablation}
	\scriptsize\setlength{\tabcolsep}{2.5pt}
	\begin{tabular}{@{}lcccccc@{}}
	\toprule
	 & \multicolumn{2}{c}{16 fragments} & \multicolumn{4}{c}{26 fragments} \\
	\cmidrule(lr){2-3}\cmidrule(l){4-7}
	 & \multicolumn{2}{c}{128\,px, stride 3} & \multicolumn{2}{c}{128\,px} & \multicolumn{2}{c}{384\,px} \\
	\cmidrule(lr){2-3}\cmidrule(lr){4-5}\cmidrule(l){6-7}
	Classifier & 10\% rule & Any contact & Stride 3 & Stride 1 & Stride 3 & Stride 1 \\
	\midrule
	ResNet18 & 0.915 & 0.758 & 0.789 & 0.786 & 0.958 & 0.955 \\
	ResNet18 + ImageNet & 0.982 & 0.846 & 0.913 & 0.914 & 0.975 & 0.971 \\
	ResNet18 + roll & 0.988 & 0.868 & 0.943 & 0.936 & 0.975 & 0.973 \\
	U-Net & 0.997 & 0.878 & 0.970 & 0.970 & 0.964 & 0.966 \\
	\bottomrule
	\end{tabular}
\end{table}
We also performed two ablations of the classifier on the \emph{Original} design with the $C_8$ model (Table \ref{tab:ablation}). When every contact is labelled as adjacent instead of applying the 10\% rule, every classifier loses between 0.119 and 0.157 of accuracy on the 16-fragment images. A ResNet18 trained with the roll augmentation labels as adjacent only 23\% of the pairs sharing less than 10\% of their perimeter, against 95\% of those sharing 20\% or more. Regarding the stride, filling the adjacency matrix at every contour point rather than every third point changes the accuracy of every classifier by at most 0.007 on the 26-fragment images, at either input size. On the other hand, a larger input does help the ResNet18 classifiers, mostly the plain one (0.789 to 0.958 at stride 3), but not the U-Net (0.970 to 0.964). Hence, the weakness of the plain ResNet18 seems to be related to the resolution of its input rather than to a lack of information in the adjacency matrix.

\subsection{Zero-shot transfer to the real dataset}
\label{sec:transfer}
\begin{table}[t]
	\centering\small
	\caption{Zero-shot transfer of the pipelines trained on synthetic data to the real dataset of PairingNet \cite{Zhou2024}: adjacency AUC of the U-Net (maximum of the band map) on 199 annotated pairs against 199 pairs of fragments from different source images, with the contours rescaled by 0.8. Mean over three seeds, standard deviation at most 0.057 with the tangent descriptors and 0.088 with the $C_8$ model. Bold: best of the table for each number of fragments.}
	\label{tab:transfer}
	\footnotesize\setlength{\tabcolsep}{5pt}
	\begin{tabular}{@{}llcc@{}}
	\toprule
	& & \multicolumn{2}{c}{AUC} \\
	\cmidrule(l){3-4}
	Design & Descriptor & 16-fragment & 26-fragment \\
	\midrule
	\multirow{3}{*}{\emph{Original}} & $C_8$ model & 0.664 & 0.718 \\
	 & LLR descriptor & 0.797 & 0.771 \\
	 & CNN descriptor & 0.694 & 0.811 \\
	\midrule
	\multirow{3}{*}{\emph{Frame-free}} & $C_8$ model & 0.694 & 0.717 \\
	 & LLR descriptor & 0.756 & 0.717 \\
	 & CNN descriptor & 0.851 & 0.852 \\
	\midrule
	\multirow{3}{*}{\emph{Frame-adjacent}} & $C_8$ model & 0.521 & 0.551 \\
	 & LLR descriptor & 0.773 & 0.751 \\
	 & CNN descriptor & 0.892 & 0.910 \\
	\midrule
	\multirow{3}{*}{\emph{Varied cuts}} & $C_8$ model & 0.663 & 0.749 \\
	 & LLR descriptor & 0.932 & 0.912 \\
	 & CNN descriptor & \best{0.943} & \best{0.929} \\
	\bottomrule
	\end{tabular}
\end{table}

None of the pipelines of this section has been trained on scanned fragments. Table \ref{tab:transfer} gives the results of each pipeline, applied without modification, on the real dataset of PairingNet \cite{Zhou2024}. It shows that the $C_8$ pipelines transfer poorly whatever their training design, with AUCs between 0.521 and 0.749, the best value being that of the 26-fragment \emph{Varied cuts} model. The LLR pipelines transfer well only when their classifier is trained on the \emph{Varied cuts} design, with AUCs of 0.932 and 0.912, against AUCs between 0.717 and 0.797 for the three other designs. Since the LLR descriptor is the same in every row and the alignment score gives an AUC of 0.921 on the same adjacency matrices, what fails to transfer in the other designs is the U-Net itself. Trained on one family of synthetic boundaries, it does not recognise the bands of real boundaries unless the training cuts have varied statistics. The CNN pipelines also transfer best from the \emph{Varied cuts} design, which gives the best AUCs of the table, 0.943 and 0.929, with standard deviations of 0.005 and 0.018 over the three seeds, and the three other designs stay between 0.694 and 0.910. The CNN descriptor transfers better than the LLR descriptor on the \emph{Frame-adjacent} and \emph{Frame-free} designs, by 0.095 to 0.159 in AUC, but not consistently on the \emph{Original} design. With the CNN descriptor, the alignment score gives an AUC of 0.944 on the adjacency matrices of the \emph{Varied cuts} pipeline, equal to the 0.943 of its U-Net within the spread of the seeds. The same holds when every pair of fragments from different source images is scored (0.921 against 0.933 and 0.940 against 0.938). Thus, on the real dataset, the U-Net is above the alignment score only with the LLR descriptor at 16 fragments and equal to it with the CNN descriptor.

\subsection{Pair-searching on PairingNet}
\label{sec:pairingnet}
\begin{table*}[t]
	\centering\small
	\begin{threeparttable}
	\caption{Pair-searching under the protocol of PairingNet on its real dataset and on its generated test split. The CNN descriptor is trained on the training split of PairingNet, mean over three seeds (standard deviation at most 0.017). The values of PairingNet, JigsawNet and the rule-based method are those of \cite{Zhou2024}. Bold: best of the dataset. Underlined: best contour-only method.}
	\label{tab:pairingnet}
	\footnotesize\setlength{\tabcolsep}{2.4pt}
	\begin{tabular}{@{}llccccc@{}}
	\toprule Dataset & Method & \Rat{5} & \Rat{10} & NDCG@5 & NDCG@10 & Input \\
	\midrule
	\multirow{5}{*}{Real} & CNN descriptor $+$ alignment score (ours) & \best{\bestc{0.794}} & \best{\bestc{0.823}} & \best{\bestc{0.708}} & \best{\bestc{0.717}} & Contour \\
	 & LLR descriptor $+$ alignment score (ours) & 0.688 & 0.748 & 0.627 & 0.649 & Contour \\
	 & PairingNet \cite{Zhou2024} & 0.415 & 0.564 & 0.321 & 0.369 & Contour + texture \\
	 & JigsawNet \cite{Le2019} & 0.237 & 0.326 & 0.185 & 0.213 & Image \\
	 & Rule-based \cite{Zhang2014} & 0.138 & 0.247 & 0.079 & 0.114 & Contour + colour \\
	\midrule
	\multirow{7}{*}{Generated} & CNN descriptor $+$ alignment score (ours) & \bestc{0.371} & \bestc{0.417} & \bestc{0.369} & \bestc{0.385} & Contour \\
	 & LLR descriptor $+$ alignment score (ours) & 0.232 & 0.266 & 0.228 & 0.241 & Contour \\
	 & PairingNet \cite{Zhou2024} & \best{0.493} & \best{0.608} & \best{0.417} & \best{0.458} & Contour + texture \\
	 & PairingNet, texture only \cite{Zhou2024} & 0.479 & 0.597 & 0.394 & 0.436 & Texture \\
	 & PairingNet, contour only \cite{Zhou2024} & 0.016 & 0.025 & 0.013 & 0.016 & Contour \\
	 & JigsawNet \cite{Le2019} & 0.380 & 0.472 & 0.326 & 0.359 & Image \\
	 & Rule-based \cite{Zhang2014} & 0.317 & 0.416 & 0.248 & 0.284 & Contour + colour \\
	\bottomrule
	\end{tabular}
	\end{threeparttable}
\end{table*}

\begin{figure*}[t]
	\centering
	\includegraphics[width=0.78\linewidth]{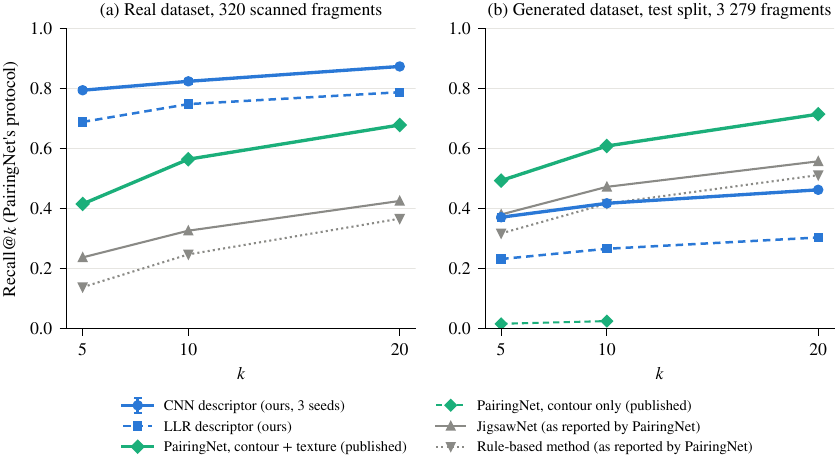}
	\caption{Pair-searching under the protocol of PairingNet. Recall@$k$ on (a) the real dataset and (b) the generated test split, for the CNN descriptor (three seeds), the LLR descriptor, and the published results of \cite{Zhou2024}. The contour-only variant of PairingNet is published for the generated dataset only, at $k=5$ and $k=10$.}
	\label{fig:pairingnet}
\end{figure*}

Table \ref{tab:pairingnet} and Figure \ref{fig:pairingnet} give the results of pair-searching under the protocol of PairingNet \cite{Zhou2024}. On the real dataset, the CNN descriptor with the alignment score reaches Recall@5, Recall@10 and Recall@20 of 0.794, 0.823 and 0.873 and an NDCG@10 of 0.717, with a standard deviation of at most 0.017 over three seeds. PairingNet, which uses the texture in addition to the contour and was trained on its own generated dataset, reaches 0.415, 0.564, 0.678 and 0.369. The LLR descriptor with the alignment score reaches a Recall@10 of 0.748. The AUC of the annotated pairs against all pairs of fragments from different source images is 0.970 for the CNN descriptor and 0.932 for the LLR descriptor. Thus, learning the comparison brings an improvement of 0.075 in Recall@10 over the LLR descriptor with the same representation and the same score. We note that the real dataset is small and favourable to geometric methods, since the edges are clean and complementary, so we do not extrapolate these results to eroded material.

On the generated test split, the ranking is reversed. The CNN descriptor reaches 0.371, 0.417 and 0.462 and an NDCG@10 of 0.385, with an AUC of 0.868 against 0.779 for the LLR descriptor. This is below PairingNet (0.493, 0.608 and 0.714), its texture-only variant (Recall@10 of 0.597) and JigsawNet (0.472), level with the rule-based method (0.416) and well above the contour-only variant of PairingNet (0.025) and the LLR descriptor (0.266). The generated cuts of PairingNet are smooth and the adjacent pairs often share short boundaries. For the LLR descriptor, the Recall@10 is 0.011, 0.202 and 0.578 on the three terciles of shared-boundary length (22 to 75, 75 to 116 and 116 to 313 points). The gap of 0.191 in Recall@10 to the full PairingNet model is the share of adjacent pairs that the contour alone does not retrieve on this split. Re-ranking the best candidates of the CNN descriptor with the alignment score or with the U-Net on the adjacency matrices of the pipelines does not improve the Recall@10 on either dataset, so no re-ranking is reported.

\subsubsection{Robustness to jitter}
\begin{figure}[t]
	\centering
	\includegraphics[width=0.9\linewidth]{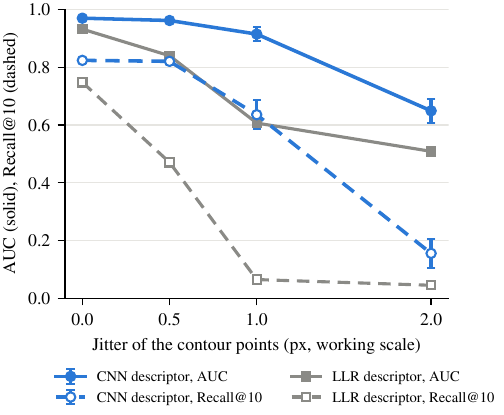}
	\caption{AUC and Recall@10 of the two descriptors on the real dataset of PairingNet under an independent Gaussian jitter of the contour points (CNN descriptor: three seeds).}
	\label{fig:jitter}
\end{figure}
Figure \ref{fig:jitter} shows the effect of an independent Gaussian jitter of the contour points of the real dataset, after rescaling. The LLR descriptor degrades quickly, since the tangent angle is the direction of a difference of two nearby contour points, on which a small jitter has a large effect, and the comparison has a fixed noise floor $\varepsilon$. Its AUC is 0.932, 0.840, 0.606 and 0.508 and its Recall@10 is 0.748, 0.470, 0.064 and 0.045 for a jitter of 0, 0.5, 1 and 2 pixels. The CNN descriptor, which was trained with a jitter of up to 0.8 pixel, keeps an AUC of 0.970, 0.962, 0.914 and 0.649 and a Recall@10 of 0.823, 0.820, 0.635 and 0.155, with a standard deviation of at most 0.050 over three seeds up to 1 pixel. Both descriptors are built on the same profile, which is consistent with the gain coming from the learned comparison and its training with jitter.

\subsubsection{Training on synthetic data}
\begin{table}[t]
	\centering\small
	\begin{threeparttable}
	\caption{Recall@10 of the CNN descriptor under the pair-searching protocol of PairingNet, according to its training dataset (rows) and test dataset (columns). Synthetic: the \emph{Varied cuts} design, 600 training images and the 100 test images (784 fragments, 735 adjacent pairs). Generated and real: the generated training split, the generated test split and the real dataset of PairingNet. Mean over three seeds (standard deviation at most 0.009, except 0.031 for the synthetic-trained model on the generated test split). Bold: best of the column.}
	\label{tab:crosstrain}
	\footnotesize\setlength{\tabcolsep}{4pt}
	\begin{tabular}{@{}lccc@{}}
	\toprule
	& \multicolumn{3}{c}{Tested on} \\
	\cmidrule(l){2-4}
	Trained on & Synthetic & Generated & Real \\
	\midrule
	Synthetic & \best{0.974} & 0.357 & \best{0.876} \\
	Synthetic, then generated & 0.945 & 0.404 & 0.825 \\
	Generated & 0.934 & \best{0.417} & 0.823 \\
	\bottomrule
	\end{tabular}
	\end{threeparttable}
\end{table}
To separate what the CNN descriptor of Table \ref{tab:pairingnet} owes to its training dataset from what it owes to its design, we also train it, with the same hyper-parameters, on the exact correspondences of 600 images of the \emph{Varied cuts} design (4\,650 fragments, 4\,329 adjacent pairs), its checkpoint being selected on 30 validation images of the same design. A third model starts from this synthetic-trained model and is fine-tuned on the training split of PairingNet for 3\,000 steps at a learning rate of $5\cdot10^{-4}$. Table \ref{tab:crosstrain} gives the Recall@10 of the three models on the test split of the \emph{Varied cuts} design, on the generated test split of PairingNet and on its real dataset, under the same protocol in every entry. Each model is the best on the dataset of its own kind: the synthetic-trained model on the synthetic test split (0.974 against 0.934) and the model trained on PairingNet on the generated test split (0.417 against 0.357). On the generated test split, the synthetic-trained model is also the least stable across seeds (between 0.322 and 0.381). On the real dataset, however, the synthetic-trained model is the best (0.876 against 0.823), although neither model has seen a scanned fragment. The fine-tuned model behaves like the model trained on PairingNet (0.945, 0.404 and 0.825) and does not combine the two advantages. The results of Table \ref{tab:pairingnet} on the real dataset therefore do not depend on training on PairingNet, and the \emph{Varied cuts} design is the better training dataset for it.

\section{Conclusion and further work}
\label{sec:conclusion}
In this paper, we kept the adjacency matrix of the two-stage pipeline of \textcite{Beaulac2023} and replaced the descriptor that fills it and the classifier that reads it. The tangent descriptor gives an accuracy at least as high as that of the $C_8$ model that it replaces, for every classifier and every design of the synthetic benchmark (Table \ref{tab:synthetic}). On the dataset of \textcite{Beaulac2023}, the LLR pipeline reaches 0.979 against 0.916 for the $C_8$ pipeline once its evaluation is corrected (Table \ref{tab:beaulac}). The U-Net is the best classifier, or tied with the best, in 21 of the 24 cases of Table \ref{tab:synthetic}, and the alignment score, which reads the same adjacency matrix without training, is the score of every pair-searching result. Trained on the \emph{Varied cuts} design, the resulting pipeline transfers to the real dataset of PairingNet with an AUC of 0.932 with the LLR descriptor and of 0.943 with the CNN descriptor (Table \ref{tab:transfer}). Under the pair-searching protocol of PairingNet (Table \ref{tab:pairingnet}), the CNN descriptor outperforms the published model, which also uses texture, on the real dataset (Recall@10 of 0.823 against 0.564), whereas the methods that use texture remain ahead on the generated test split. Moreover, the CNN descriptor trained on synthetic correspondences alone reaches a Recall@10 of 0.876 on the real dataset (Table \ref{tab:crosstrain}) and, under a jitter of one pixel on the contour points, it keeps a Recall@10 of 0.635 against 0.064 for the LLR descriptor (Figure \ref{fig:jitter}).

Some questions remain open. First, the CNN descriptor is trained on point correspondences alone and the U-Net is trained afterwards on its adjacency matrices. Training the two jointly could make a single pipeline serve both adjacency prediction and pair-searching, whereas fine-tuning the descriptor alone on the training split of PairingNet did not improve it on the real dataset (Table \ref{tab:crosstrain}). Second, texture information could be added as supplementary channels of the window descriptor, since texture is what the full PairingNet model adds on the generated dataset, and the structure of the adjacency matrix does not depend on what fills its entries. Finally, the eroded fragments of RePAIR \cite{Tsesmelis2024} would provide a test that neither our synthetic benchmark nor the clean scans of PairingNet provide.

\paragraph{Limitations.}
Two limitations should be kept in mind. First, the accuracies of Tables \ref{tab:beaulac} and \ref{tab:ablation} are measured on the validation split, which is also the split on which the best epoch of each classifier is selected, so these values are slightly optimistic. In Table \ref{tab:synthetic}, the seeds of the $C_8$ pipelines differ by up to 0.198 in accuracy, against 0.044 for the LLR pipelines, so the individual $C_8$ entries should be read with this margin. Second, the parameters of the LLR descriptor were chosen on the datasets of PairingNet, partly on the real dataset (Appendix \ref{sec:provenance}), so its real-dataset results are not those of a blind run, whereas those of the CNN descriptor are.

\paragraph{LLM usage statement.} A large language model was used as a writing and programming assistant. The authors verified every statement, number and script.

{\footnotesize
\printbibliography
}

\appendix
\section{Provenance of the LLR parameters}
\label{sec:provenance}
The LLR descriptor has five parameters, namely the window, the spacing, the smoothing, the noise floor $\varepsilon$ and the clip value. A first version of its score was inspected on a sample of 80 true and 320 false pairs of the real dataset of PairingNet \cite{Zhou2024}. The noise floor was then measured on 60 true real alignments (median of $D$ equal to 0.0024, and $D\approx I$ on random alignments), which motivated the likelihood ratio of Equation \eqref{eq:W} and the value $\varepsilon=0.003$. Eleven parameter settings were then compared on 100 true and 400 false pairs of the \emph{training} split of PairingNet, where a setting of 18 samples every 2 points with $\sigma=2$ was selected. The faster setting used in the paper ($h=4$, 9 samples every 4 points) was the best of three on the real dataset. Hence, the real-dataset values of the LLR rows of Tables \ref{tab:transfer} and \ref{tab:pairingnet} are not those of a blind run, while their values on the generated dataset and every value of the CNN descriptor are. The hyper-parameters of the CNN descriptor (Section \ref{sec:descriptor}) were fixed once on the training split of PairingNet and its checkpoint is the best of four evaluations on the validation split (per-query validation Recall@10 0.502, 0.508 and 0.523 for the three seeds).

\section[Held-out splits on the dataset of Beaulac]{Held-out splits on the dataset of \citeauthor{Beaulac2023}}
\label{sec:splits}
A held-out split must be chosen with care on the dataset of \textcite{Beaulac2023}, who released a 50-image set and a 100-image set, each with its own validation images. The validation split of the 50-image set consists of training images of the 100-image set used in this paper: each of its 375 adjacent pairs comes from one of the training images 51 to 65. Hence, a classifier trained on the 100-image set and scored on the validation split of the 50-image set would be scored on its own training images. For this reason, Table \ref{tab:beaulac} reports the validation split of the 100-image set and no held-out test value.

\section[Larger neighbourhoods on the dataset of Beaulac]{Larger neighbourhoods on the dataset of \citeauthor{Beaulac2023}}
\label{sec:beaulac63}
The code of \textcite{Beaulac2023} was also run with neighbourhoods of 63 pixels and a stride of 5, which combines two settings of \cite{Beaulac2023} (reported accuracies of 0.709 for neighbourhoods of 63 pixels and 0.714 for a stride of 5). With the roll augmentation, the $C_8$ model gives a validation accuracy of 0.864 and the classifier gives 0.700 with the labels as written and 0.902 with the corrected labels.

\section{Hyper-parameters}
\label{sec:hyper}
\emph{Generator}: five octaves of one-dimensional noise, relative amplitude 0.05, initial lacunarity 8, amplitude gain 0.5, lacunarity gain 2, side bound 0.05 (16 fragments) or 0.03 (26), $\beta=0.1$. \emph{Varied cuts}: amplitude $U(0.02,0.10)$, lacunarity in $\{4,6,8,12,16\}$, amplitude gain $U(0.35,0.7)$, lacunarity gain $U(1.6,2.5)$, 3--6 octaves, smoothstep interpolation with probability 0.5, deviation from the chord up to 0.15 of its length, end-point bound $U(0.03,0.15)$. \emph{$C_8$ model}: fields 24-48-48-96-96-64, hidden 64, neighbourhoods of 29 pixels restricted to the inscribed disc, $\delta=3$ pixels, random rotation and swap, Adam $5\cdot10^{-4}$, weight decay $10^{-5}$, batches of 32, 15 epochs, 2\,400 training pairs, best epoch on 4\,000 rotated validation pairs. \emph{LLR descriptor}: 9 samples every 4 points, $\sigma=2$, $\varepsilon=0.003$, clip $-2$. \emph{CNN descriptor}: $\sigma\in\{1,2,4\}$, $h=10$, $\Delta=2$, width 64, dimension 32, entries every 4 points, clip $-2$, jitter up to 0.8 pixel in 4 copies, batches of 256, 12\,000 steps, AdamW with a learning rate of $2\cdot10^{-3}$, a one-cycle schedule and a weight decay of $10^{-4}$, hard negatives at $\ge24$ points. \emph{Classifier}: stride 3, 2\,400/800 adjacency matrices, 128 pixels (224 on the dataset of \textcite{Beaulac2023}), ResNet18 with Adam $10^{-3}$ and batches of 64, U-Net (16-32-64 channels) with Adam $10^{-3}$ and batches of 32, positive weight 20 and Dice on the band, roll augmentation, 30 epochs, best validation epoch. \emph{PairingNet contours}: rescaling factor 0.4 (descriptors) or 0.8 (pipelines trained on 1000-pixel images).

\end{document}